\documentclass{article}
\usepackage[utf8]{inputenc}
\usepackage[margin = 1in]{geometry}
\usepackage{authblk}

\usepackage{amsmath, amssymb, amsthm, amsfonts, mathtools, multirow, mathrsfs, tabu, booktabs, lipsum, url, subcaption, graphicx,xcolor, tabu,  authblk, hyperref, blindtext}

\usepackage[english]{babel}
\usepackage[shortlabels]{enumitem}
 \usepackage[ruled,vlined, linesnumbered]{algorithm2e}
\usepackage[capbesideposition=outside,capbesidesep=quad]{floatrow}
\usepackage[framemethod=tikz]{mdframed}
\usepackage{bbm}

\allowdisplaybreaks

\newcommand{\appropto}{\mathrel{\vcenter{
  \offinterlineskip\halign{\hfil$##$\cr
    \propto\cr\noalign{\kern2pt}\sim\cr\noalign{\kern-2pt}}}}}
    
\newcommand{\R}{\mathbb{R}} 

\newcommand{\N}{\mathbb{N}} 

\DeclareMathOperator*{\argmin}{argmin}

\title{Active Diffusion and VCA-Assisted Image Segmentation of
Hyperspectral Images}
\author[1]{Sam L. Polk}
\author[2]{Kangning Cui}
\author[3]{Robert J. Plemmons}
\author[1]{James M. Murphy\footnote{Corresponding Author: JM.Murphy@Tufts.edu \newline This work was partially funded by  the US National Science Foundation grants NSF-DMS 1912737, NSF-DMS 1924513, and NSF-CCF 1934553.}}
\affil[1]{ Department of Mathematics, Tufts University}
\affil[2]{ Department of Mathematics, City University of Hong Kong}
\affil[3]{ Departments of Mathematics and Computer Science, Wake Forest University}
\date{}                     %% if you don't need date to appear
\begin{document}

% \threeauthors
%  {Sam L. Polk\sthanks{Corresponding Author; email: \url{Samuel.Polk@Tufts.edu} \newline This work was partially funded by  the US National Science Foundation grants NSF-DMS 1912737, NSF-DMS 1924513, and NSF-CCF 1934553.} and James M. Murphy}
% 	{Tufts University\\
% 	Department of Mathematics\\
% 	Medford, MA, USA}
%  {Kangning Cui}
% 	{City University of Hong Kong\\
% 	Department of Mathematics\\
%     Hong Kong, Hong Kong}
%  {Robert J. Plemmons}
% 	{Wake Forest University\\
% 	Department of Mathematics\\
%     Wake Forest, NC, USA}

\topmargin=0mm

\maketitle
\begin{abstract}
Hyperspectral images   encode rich structure that can be exploited for material discrimination by machine learning algorithms. This article introduces the Active Diffusion and VCA-Assisted Image Segmentation (ADVIS) for active material discrimination. ADVIS selects high-purity, high-density pixels that are far in diffusion distance (a data-dependent metric) from other high-purity, high-density pixels in the hyperspectral image. The ground truth labels of these pixels are queried and propagated to the rest of the image. The ADVIS active learning algorithm is shown to strongly outperform its fully unsupervised clustering algorithm counterpart, suggesting that the incorporation of a very small number of carefully-selected ground truth labels can result in substantially superior material discrimination in hyperspectral images. 

% This document is a model and instructions for \LaTeX.
% This and the IEEEtran.cls file define the components of your paper [title, text, heads, etc.]. *CRITICAL: Do Not Use Symbols, Special Characters, Footnotes, 
% or Math in Paper Title or Abstract.
\end{abstract}

Semi-supervised Machine Learning.

\section{Introduction}

Hyperspectral images (HSIs) are high-dimensional remotely-sensed images that encode rich information about a scene~\cite{eismann2012hyperspectral}. Despite storing reflectance in a hundred or more spectral bands, HSIs typically encode intrinsically low-dimensional structure that can be exploited by machine learning algorithms for image segmentation~\cite{murphy2018unsupervised, DVIS, wang2017novel}. While HSIs are an important data source for material discrimination, their use for this task is complicated by at least two key factors. First, the number of pixels in typical HSIs can be very large, rendering manual labeling and analysis infeasible~\cite{eismann2012hyperspectral}. Second, the spatial resolution of HSIs is often coarse, so any one pixel may correspond to a spatial region that contains multiple materials~\cite{DVIS}. Thus, efficient machine learning algorithms that rely on few expert annotations or labels are needed to capture the latent material structure in HSIs. 

This article introduces the Active Diffusion and VCA-Assisted Image Segmentation (ADVIS) algorithm for material discrimination in HSIs using Vertex Component Analysis (VCA)~\cite{nascimento2005VCA}. ADVIS is an \emph{active learning} algorithm based on the unsupervised Diffusion and VCA-Assisted Image Segmentation (D-VIS) clustering algorithm. D-VIS is closely related to Diffusion and Volume maximization-based Image Clustering (D-VIC), which has been shown to perform well at material discrimination on benchmark HSIs~\cite{DVIS}. We show that incorporating just a few carefully-chosen expert labels in ADVIS substantially improves algorithm performance. 

The rest of this article is structured as follows. In Section \ref{sec: background}, background is provided on HSI segmentation, diffusion geometry, spectral unmixing, and D-VIS. In Section \ref{sec: ADVIS}, the ADVIS algorithm for active material discrimination is introduced. Section \ref{sec: numerical experiments} contains numerical experiments where ADVIS is compared against D-VIS on real HSI data. In Section \ref{sec: conclusion}, we conclude and discuss future work.

\section{Background} \label{sec: background}

\subsection{Hyperspectral Image Segmentation} \label{sec: HSIS}

An HSI segmentation algorithm partitions pixels of an HSI $X=\{x_i\}_{i=1}^n \subset~\R^D$ (interpreted as a point cloud, where $n$ is the number of pixels and $D$ is the number of spectral bands) into groups $\{X_k\}_{k=1}^K$ sharing key commonalities (e.g., common materials)~\cite{friedman2001elements}. Unsupervised HSI segmentation (also called \emph{clustering}) algorithms do not rely on ground truth labels to obtain the partition $\{X_k\}_{k=1}^K$. In contrast, \emph{semi-supervised} and \emph{active learning} HSI segmentation algorithms rely on the ground truth labels of a few pixels to partition $X$.

\subsection{Diffusion Geometry} \label{sec: diffusion geometry}

To exploit the intrinsic low-dimensionality of HSIs, graph-based HSI segmentation algorithms identify HSI pixels as nodes in an undirected graph~\cite{coifman2006diffusion}. Edges between pixels are encoded in an adjacency matrix $\textbf{W}\in\R^{n\times n}$, where $\textbf{W}_{ij}=1$ if $x_j$ is one of the $N$ nearest neighbors of $x_i$ and $\textbf{W}_{ij}=0$ otherwise. Define $\textbf{P} = \textbf{D}^{-1}\textbf{W}$, where $\textbf{D}$ is the diagonal matrix with $\textbf{D}_{ii} = \sum_{j=1}^n \textbf{W}_{ij}$.  The matrix $\textbf{P}$ can be identified as the transition matrix for a Markov diffusion process on HSI pixels. Assuming $\textbf{P}$ is reversible, aperiodic, and irreducible,  there is a unique $\pi\in\R^{1\times n}$ satisfying $\pi \textbf{P}=\pi$. 

\emph{Diffusion distances} enable  direct comparisons between pixels in the context of the diffusion process encoded in $\textbf{P}$~\cite{coifman2006diffusion}. Define the diffusion distance at time $t\geq 0$ between pixels $x_i,x_j\in X$ by
\[D_t(x_i, x_j) = \sqrt{\sum_{k=1}^n \frac{\left[\left(\textbf{P}^t\right)_{ik}-\left(\textbf{P}^t\right)_{jk}\right]^2}{\pi_k }}.\]
For datasets with well-separated and highly coherent classes, the within-class diffusion distance is bounded away from the between-class diffusion distance across a broad range of $t$~\cite{murphy2021multiscale}. Thus, diffusion distances are a useful tool for HSI segmentation.  Diffusion distances can be related to the eigendecomposition of $\textbf{P}$ via 
\[D_t(x_i,x_j) = \sqrt{\sum_{k=1}^n|\lambda_k|^{2t}\left[\left(\psi_k\right)_i -\left(\psi_k\right)_j\right]^2},\] 
where $\{(\lambda_k, \psi_k)\}_{k=1}^n$ are the right eigenvalue-eigenvector pairs of $\textbf{P}$~\cite{coifman2006diffusion}. For $t$ sufficiently large, eigenvectors with $|\lambda_k|^t\approx 0$ can be discarded, yielding a low-cost, accurate approximation of diffusion distances.  

\subsection{Spectral Unmixing} \label{sec: spectral unmixing}

HSIs are often recorded at a coarse spatial resolution, so a single pixel may correspond to a spatial region that contains multiple materials~\cite{ bioucas2008HySime, chan2011simplex, nascimento2005VCA}. \emph{Spectral unmixing} algorithms may be used to estimate the proportions of materials within each pixel~\cite{ chan2011simplex, nascimento2005VCA}. Mathematically, if $m$ is the number of materials in the scene, linear spectral unmixing algorithms learn two matrices, $\textbf{A}\in\R^{n\times m}$ (called abundances) and $\textbf{U}=~(u_1 \; u_2\;\dots \; u_m)^\top\in \R^{m\times D}$ (called endmembers) such that $x_i \approx \sum_{j=1}^m \textbf{A}_{ij}u_j$ for each $x_i\in X$. Each $u_i$ is the intrinsic spectral signature of a material, and the rows of $\textbf{A}$ encode the relative abundances of materials in HSI pixels~\cite{ chan2011simplex, nascimento2005VCA}. The \emph{purity} of the pixel $x_i$, defined by $\eta(x_i) = \max_{1\leq j\leq m} \textbf{A}_{ij}$, will thus be large for pixels that predominantly contain just one material and small elsewhere~\cite{DVIS}.

\subsection{Diffusion and VCA-Assisted Image Segmentation} \label{sec: D-VIS}

D-VIS (Algorithm \ref{alg: D-VIS}) is an unsupervised material discrimination algorithm, meaning no expert labels are used to obtain an HSI segmentation. D-VIS operates in two main stages. In the first, D-VIS learns an estimate for pixels that are exemplary of all underlying material classes (called \emph{class modes}) and assigns these pixels unique labels.  D-VIS propagates the labels of class modes to non-modal pixels in its second stage.  

D-VIS first performs spectral unmixing of $X$ to calculate $\eta(x)$, using HySime to learn $m$~\cite{bioucas2008HySime} and VCA to learn endmembers~\cite{ bro1997fast, nascimento2005VCA}. This differs slightly from D-VIC, which relies on Alternating Volume Maximization to learn endmembers~\cite{ chan2011simplex, DVIS}. Next, D-VIS calculates empirical density: 
\[p(x) =~\sum_{y\in NN_N(x)}\exp\left(-\frac{\|x-y\|_2^2}{\sigma_0^2}\right),\]
where $NN_N(x)$ is the set of $N$ nearest neighbors of $x$ in $X$ and $\sigma_0>0$ is a \emph{density scale} that controls the interaction radius between pixels. D-VIS incorporates pixel purity and empirical data density into a single measure of pixel quality 
\[\zeta(x) = \frac{2\Bar{p}(x) \Bar{\eta}(x)}{\Bar{p}(x)+ \Bar{\eta}(x)},\]
where $\Bar{p}(x) = \frac{p(x)}{\max_{y\in X}p(y)}$ and $\Bar{\eta}(x) = \frac{\eta(x)}{\max_{y\in X}\eta(y)}$. Thus, $\zeta(x)$ is the harmonic mean of pixel purity and density, normalized so that each is on the same scale.  Note $\zeta(x)$ will be large for pixels that are modal (with high $p$-value) and representative of a single material class (with high $\eta$-value). 

The second main function used for mode detection is
\begin{align*}
    d_t(x) = 
    \begin{cases} 
        \max\limits_{y\in X}D_t(x,y) & x = \argmin\limits_{y\in X}\zeta(y),\\
        \min\limits_{y\in X}\{D_t(x,y)| \zeta(y)\geq \zeta(x)\} & \text{otherwise,} 
    \end{cases}
\end{align*}
which returns the diffusion distance at time $t$ between $x$ and its $D_t$-nearest neighbor of higher density and purity for pixels that are not $\zeta$-maximizers. Maximizers of $\mathcal{D}_t(x) = \zeta(x) d_t(x)$ are high-density, high-purity pixels that are far in diffusion distance at time $t$ from other high-density, high-purity pixels, making them reasonable choices as exemplars for underlying material class structure. The $K$ maximizers of  $\mathcal{D}_t(x)$ are assigned unique labels and are treated as class modes.  Non-modal labels are assigned in order of non-increasing $\zeta(x)$ according to the label of their $D_t$-nearest neighbor that is already labeled and has  a higher $\zeta$-value.  

\begin{algorithm}[t]
\SetAlgoLined
 \KwIn{ $X$ (HSI), $N$ (\# neighbors), $K$ (\# classes)  $\sigma_0$ (density scale),   $t$ (diffusion time),}
\KwOut{$\hat{\mathcal{C}}$ (HSI segmentation)}
Compute $\eta(x)$, using HySime~\cite{bioucas2008HySime} to estimate $m$ and VCA~\cite{bro1997fast, nascimento2005VCA} for spectral unmixing\;
For each $x\in X$, compute $\zeta(x) = \frac{2\Bar{p}(x) \Bar{\eta}(x)}{\Bar{p}(x)+ \Bar{\eta}(x)}$\;
Sort $X$ according to $\mathcal{D}_t(x) = \zeta(x) d_t(x)$ in non-increasing order. Denote this sorting $\{x_{m_k}\}_{k=1}^n$. Label $\hat{\mathcal{C}}(x_{m_k}) = k$ for $1\leq k \leq K$\;
In order of non-increasing $\zeta(x)$, for each unlabeled $x\in X$, assign the label $\hat{\mathcal{C}}(x) = \hat{\mathcal{C}}(x^*)$, where $x^* = \argmin\limits_{y\in X}\{D_t(x,y)|\zeta(y)\geq \zeta(x) \ \land \ \hat{\mathcal{C}}(y)>0\}$\;
\caption{Diffusion and VCA-Assisted Image Segmentation (D-VIS)}\label{alg: D-VIS}
\end{algorithm}

\section{Active Diffusion and VCA-Assisted Image Segmentation} \label{sec: ADVIS}

Though HSI segmentation can be performed without the aid of ground truth labels, incorporating the labels of just a few carefully-chosen pixels may significantly improve the predictive capacity of an HSI segmentation algorithm. \emph{Active learning} algorithms query the ground truth labels (denoted $\mathcal{C}_{GT}$) of $B\in\N$ (called the \emph{budget}) pixels. It is typically desired that these pixels exemplify underlying class structure, as queried points' labels are propagated to unlabeled pixels. It has been shown that active learning algorithms often substantially outperform their unsupervised counterparts~\cite{  haut2018active, maggioni2019LAND, murphy2020spatially, murphy2018unsupervised, tuia2011survey, wang2017novel}. 

In this section, we introduce the Active Diffusion and VCA-Assisted Image Segmentation (ADVIS) algorithm for material discrimination (see Algorithm \ref{alg: ADVIS}). ADVIS is similar to the D-VIS clustering algorithm, with a crucial difference in the manner in which class modes are labeled. ADVIS queries the labels  of the $B$ pixels that maximize $\mathcal{D}_t(x)$. If any classes remain unlabeled after the budget expires, ADVIS reverts to unsupervised D-VIS mode estimation.  By ensuring that class modes are correctly labeled in its first stage, the ADVIS algorithm improves all labeling downstream with computational complexity identical to that of D-VIS.

% More explicitly describe the labeling procedure
\begin{algorithm}[t]
\SetAlgoLined
 \KwIn{ $X$ (HSI), $K$ (\# classes), $N$ (\# neighbors),   $\sigma_0$ (density scale),   $t$ (diffusion time), $B$ (budget)}
\KwOut{$\hat{\mathcal{C}}$ (HSI segmentation)}
Compute $\eta(x)$, using HySime~\cite{bioucas2008HySime} to estimate $m$ and VCA~\cite{ bro1997fast, nascimento2005VCA} for spectral unmixing\;
For each $x\in X$, compute $\zeta(x) = \frac{2\Bar{p}(x) \Bar{\eta}(x)}{\Bar{p}(x)+ \Bar{\eta}(x)}$\;
Sort $X$ by $\mathcal{D}_t(x) = \zeta(x) d_t(x)$ in non-increasing order. Denote this sorting$\{x_{m_k}\}_{k=1}^n$. Assign $\hat{\mathcal{C}}(x_{m_k}) = \mathcal{C}_{GT}(x_{m_k})$  for $1\leq k \leq B$ \;
Let $I=\{i_1, i_2, \dots, i_{L}\} \subset\{1,2,\dots, K\}$ be the set of classes without a labeled point. If $I$ is nonempty, label 
$\hat{\mathcal{C}}(x_{m_{B+k}}) = i_k$ for $1\leq k \leq L$\;
In order of non-increasing $\zeta(x)$, for each unlabeled $x\in X$, assign the label $\hat{\mathcal{C}}(x) = \hat{\mathcal{C}}(x^*)$, where $x^* = \argmin\limits_{y\in X}\{D_t(x,y)|\zeta(y)\geq \zeta(x) \ \land \ \hat{\mathcal{C}}(y)>0\}$\;
\caption{Active Diffusion and VCA-Assisted Image Segmentation (ADVIS)} \label{alg: ADVIS}
\end{algorithm}

\begin{figure}[b]
    \centering
    \includegraphics[width = 0.33\textwidth]{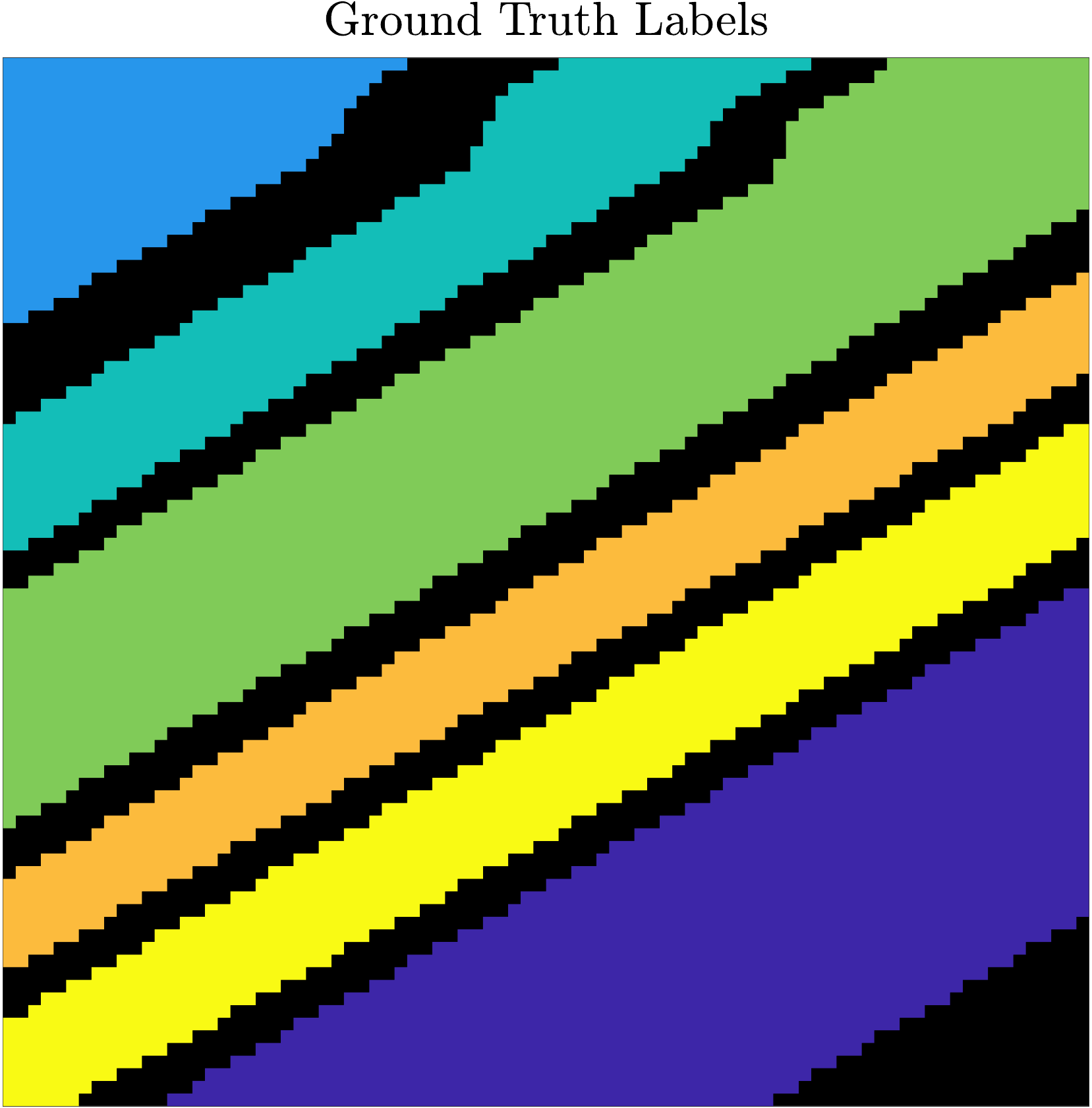} \hspace{0.1in} \includegraphics[width = 0.33\textwidth]{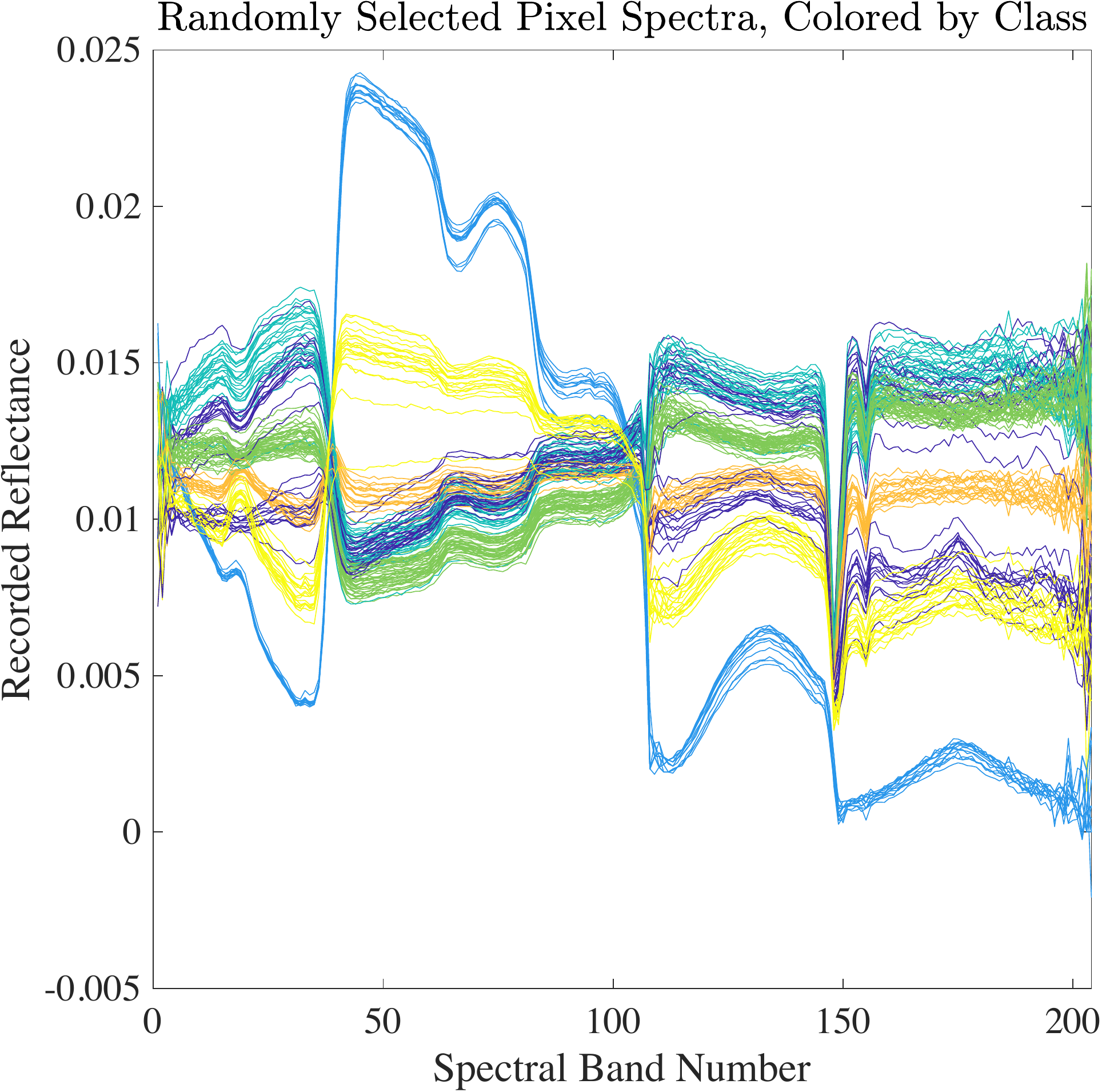}
    \caption{Ground truth labels and randomly-selected pixel spectra for the Salinas A HSI. The class indicated in purple (8-week romaine) has high intra-class spectral variability. }
    \label{fig:GT Data}
\end{figure}

\begin{figure}[t]
    \centering
    \includegraphics[width = 0.5\textwidth]{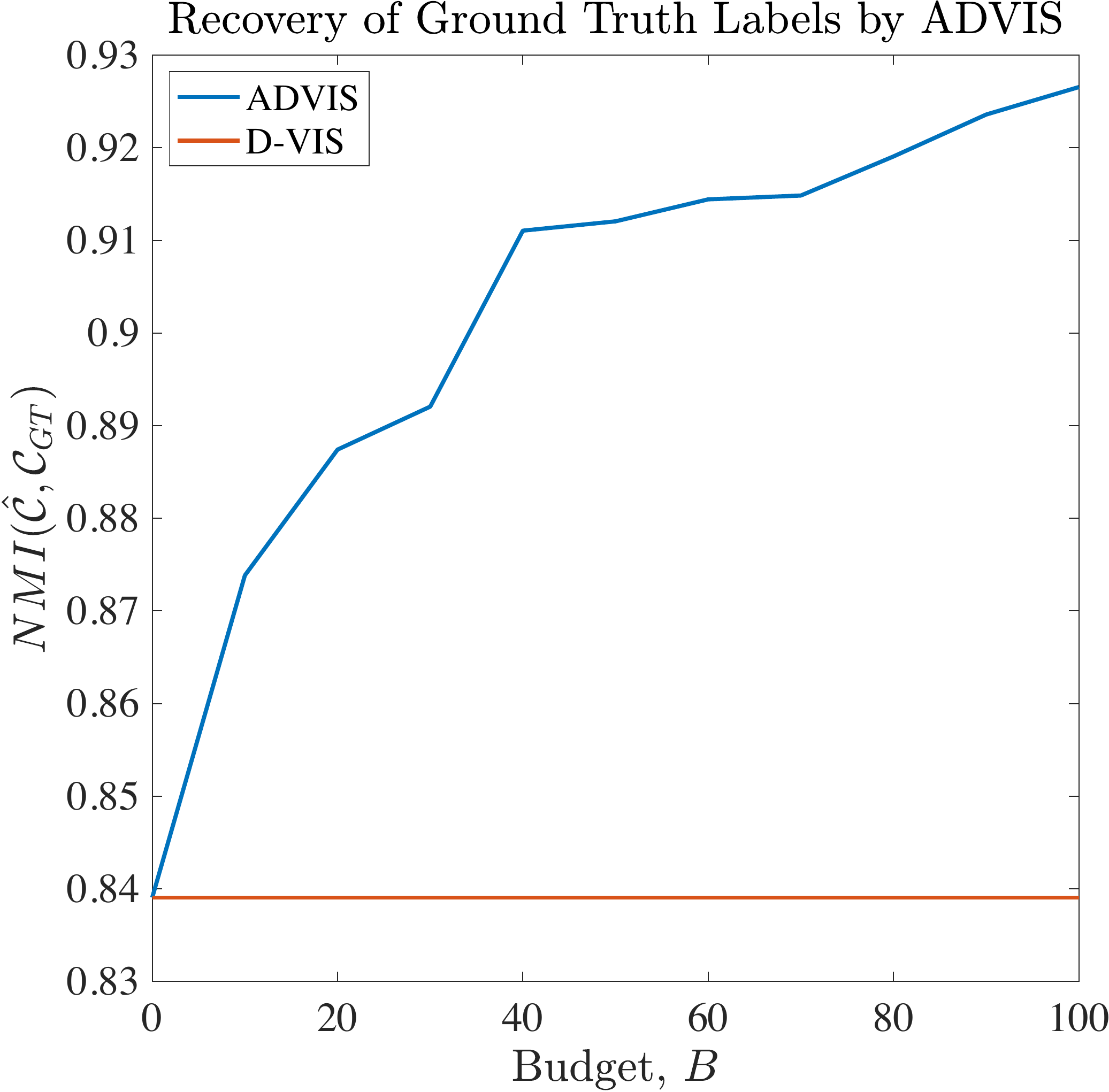}
    \caption{Comparison of the performance of D-VIS and ADVIS as a function of budget, $B$. The performance of ADVIS quickly overtakes the performance of D-VIS and monotonically increases as a function of $B$.}
    \label{fig:performance}
\end{figure}

\section{Numerical Experiments} \label{sec: numerical experiments}

This section illustrates the efficacy of ADVIS by comparing it against its unsupervised clustering counterpart (D-VIS) on the Salinas A benchmark HSI (Fig. \ref{fig:GT Data}). The Salinas A HSI was generated using the Airborne Visible/Infrared Imaging Spectrometer sensor over Salinas Valley, CA, USA and encodes $D=204$ spectral bands across $83\times 86$ pixels. D-VIS and ADVIS were evaluated on labeled pixels of the Salinas A HSI with parameters $N=320$, $\sigma_0 = 1.14\times 10^{-3}$, and $t=2^5$. For ADVIS, the budget $B$ ranged $\{10, 20, \dots, 100\}$. Performance was measured using $NMI(\hat{\mathcal{C}}, \mathcal{C}_{GT})$: the normalized mutual information between an estimated partition $\hat{\mathcal{C}}$ and the ground truth labels $\mathcal{C}_{GT}$. Before labeling, pixel purity was averaged across 100 runs to account for VCA's stochasticity.

\begin{figure}[b]
    \centering
    \includegraphics[width = 0.23\textwidth]{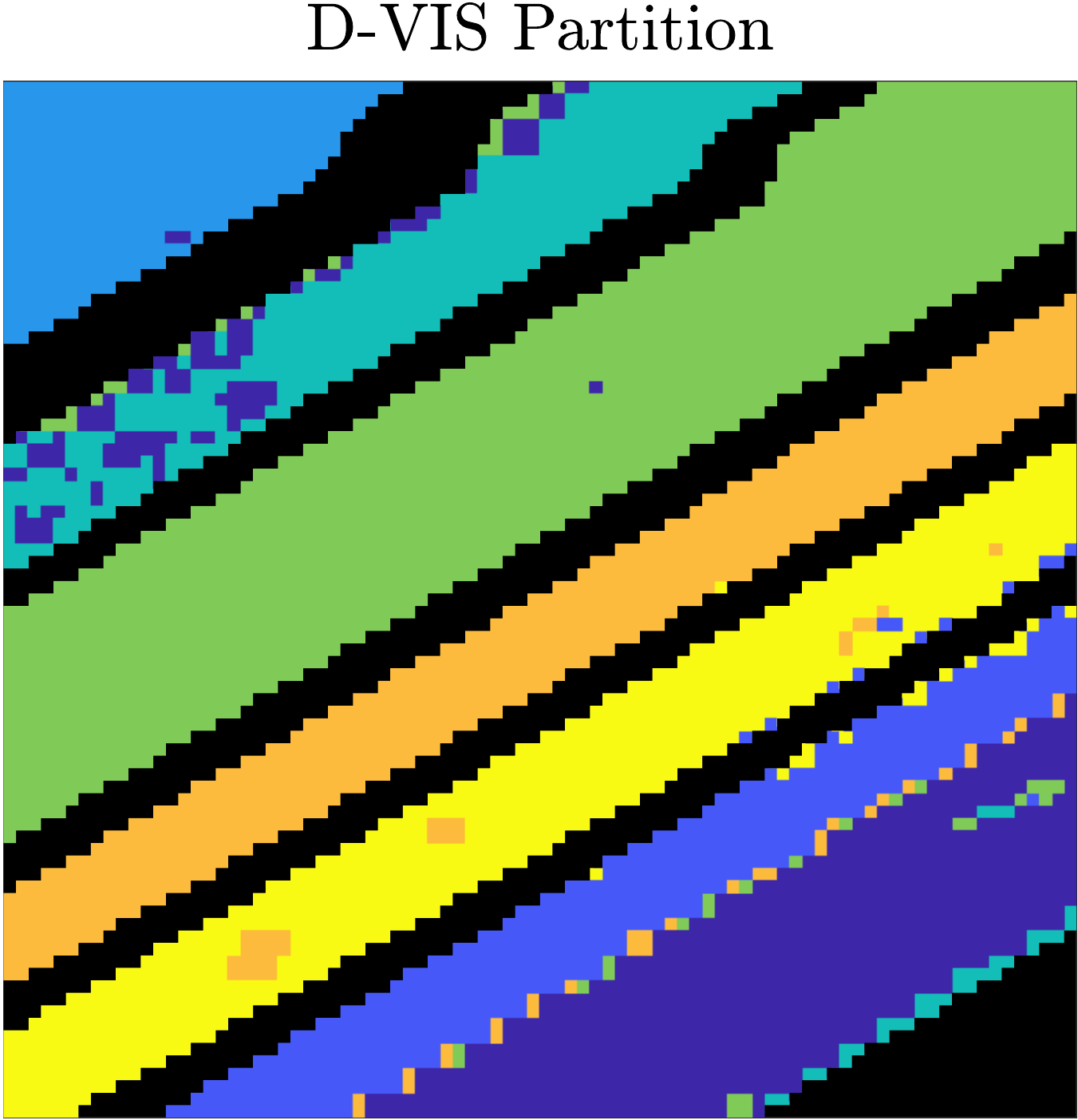} \hspace{0.05in} \includegraphics[width = 0.23\textwidth]{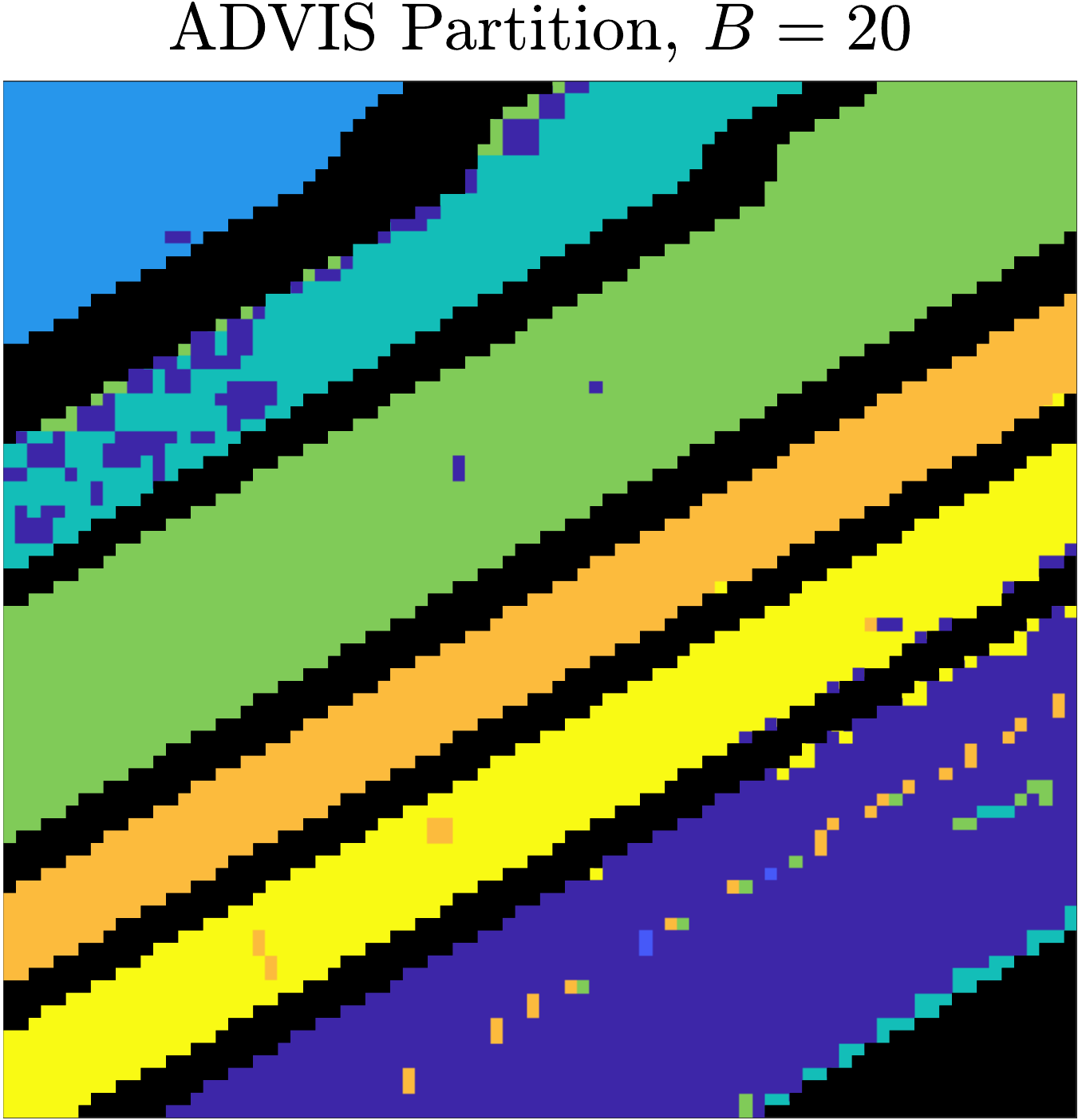}\hspace{0.05in}
    \includegraphics[width = 0.23\textwidth]{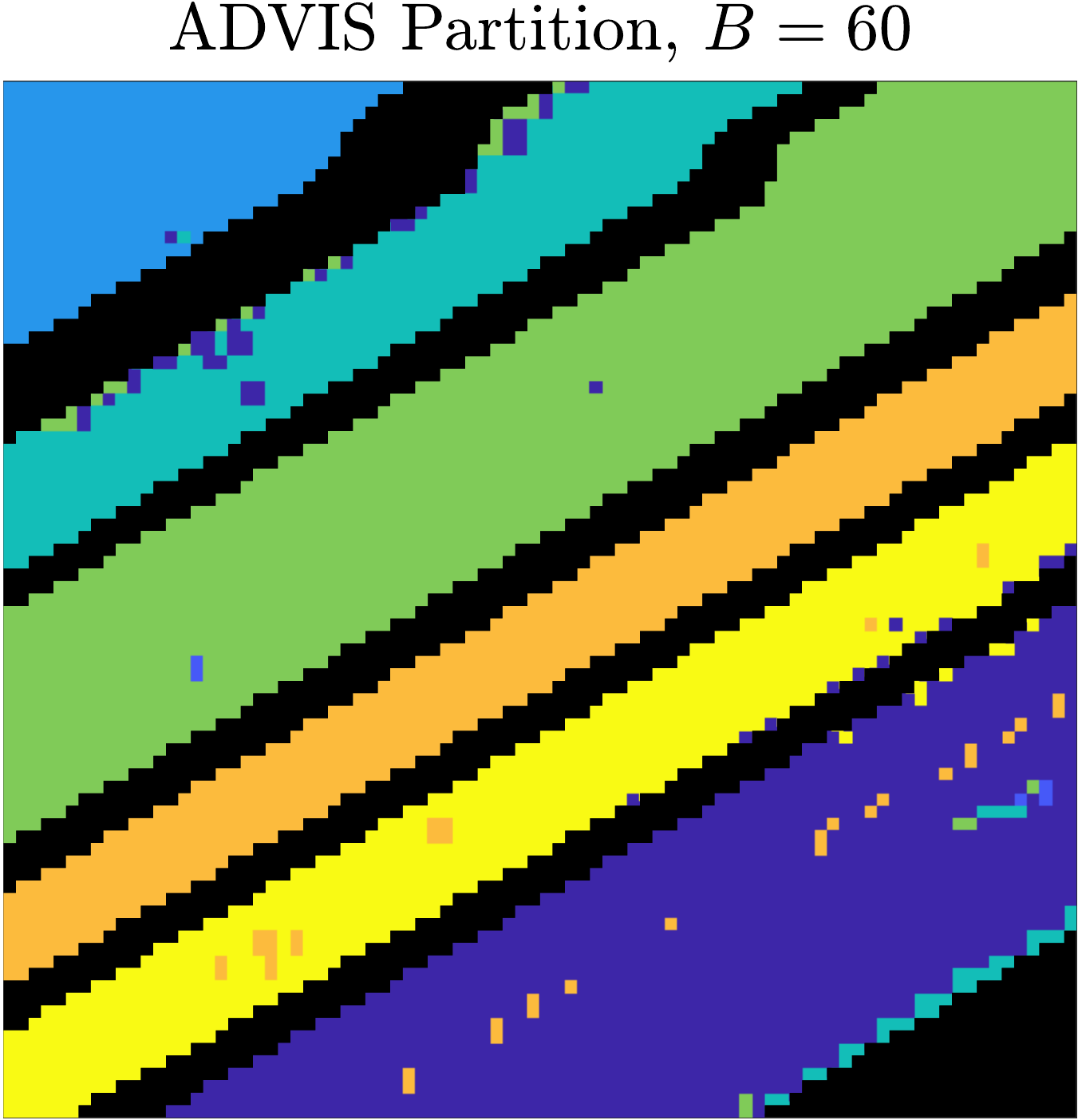} \hspace{0.05in} \includegraphics[width = 0.23\textwidth]{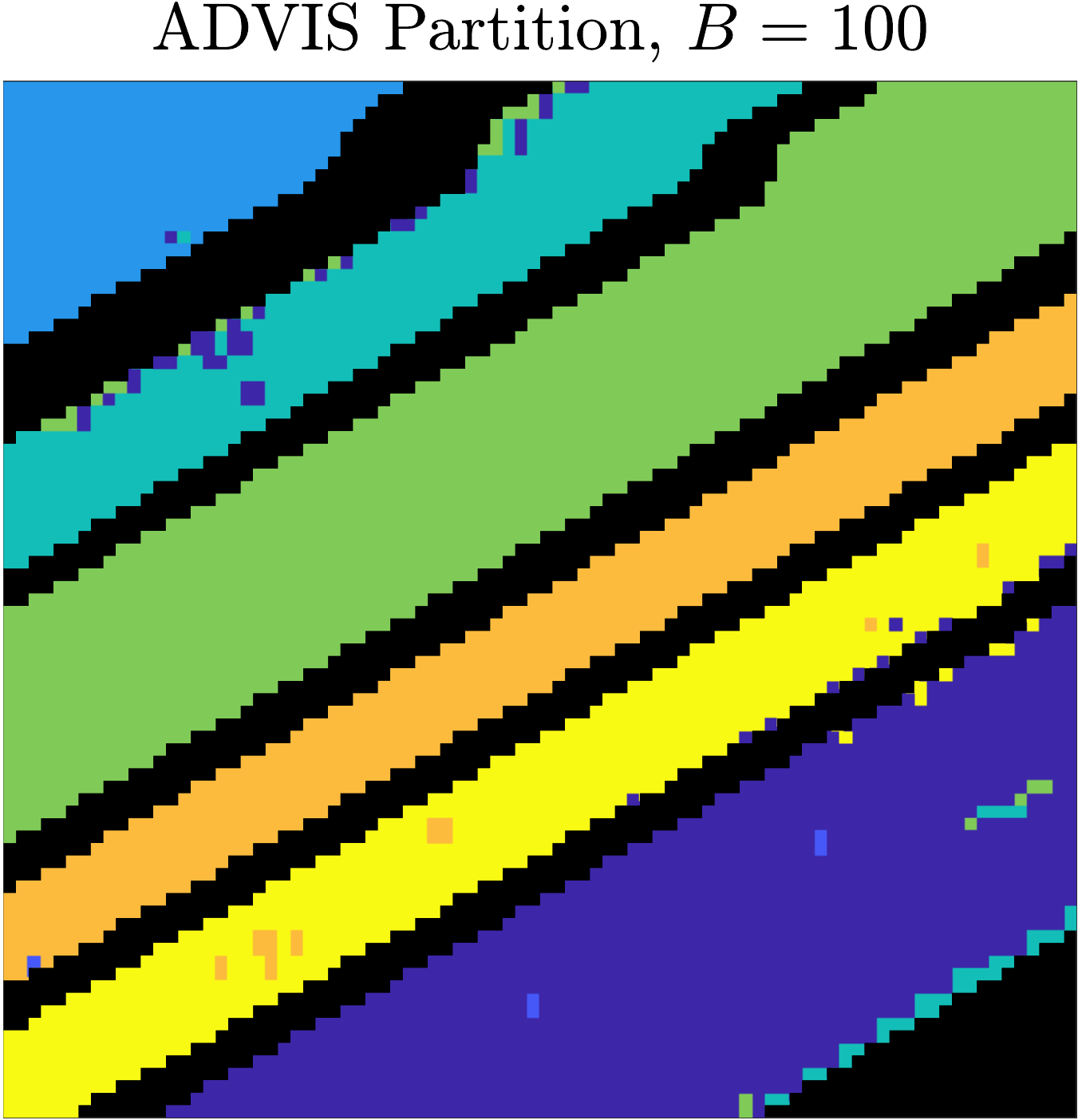}
    
    \caption{Comparison of D-VIS and ADVIS on Salinas A. The major error present in the unsupervised labeling (splitting 8-week romaine) is corrected using just $B=20$ label queries.}
    \label{fig: clusterings}
\end{figure}

In Fig. \ref{fig:performance},  the performance of ADVIS is plotted against the budget $B$. Fig. \ref{fig: clusterings} visualizes a sample of learned partitions. These results make clear that  an active learning framework with even a small budget offers a major improvement in material discrimination.  Indeed, though D-VIS erroneously splits the purple class (8-week romaine) in two, ADVIS correctly groups these pixels with just $B= 20$ ground truth labels. ADVIS labelings quickly converge to $\mathcal{C}_{GT}$ as $B$ increases, and when $B= 100$, there is little difference between the ground truth labels and the partition estimated by ADVIS. Importantly, ADVIS does not rely on spatial information, so much of the remaining error may be corrected in a spatially regularized regime~\cite{murphy2020spatially,  murphy2019spectral, polk2021multiscale}. Nevertheless, it is clear that the inclusion of a few carefully-chosen labels in ADVIS results in image segmentations of the Salinas A HSI that are substantially close to its ground truth labels.

Software to replicate numerical experiments is available on GitHub at \url{https://github.com/sampolk/D-VIC}.

\section{Conclusions} \label{sec: conclusion}

We conclude that an active learning framework that enables the incorporation of a few ground truth labels for material discrimination substantially improves the performance of an HSI segmentation algorithm. We expect that ADVIS can be extended for multiscale HSI segmentation, wherein a hierarchy of image segmentations is learned~\cite{murphy2021multiscale, polk2021multiscale}. In addition, ADVIS is likely to benefit from a modified, spatially-regularized graph, wherein edges between pixels are restricted to spatial nearest neighbors~\cite{murphy2020spatially,murphy2019spectral, polk2021multiscale}.

\bibliographystyle{siam} 
\bibliography{ref}

\begin{thebibliography}{10}

\bibitem{bioucas2008HySime}
{\sc J.~M. Bioucas-Dias and J.~M.~P. Nascimento}, {\em Hyperspectral subspace
  identification}, IEEE Trans Geosci Remote Sens, 46 (2008), pp.~2435--2445.

\bibitem{bro1997fast}
{\sc R.~Bro and S.~De~Jong}, {\em A fast non-negativity-constrained least
  squares algorithm}, J Chemom, 11 (1997), pp.~393--401.

\bibitem{chan2011simplex}
{\sc T.-H. Chan, W.-K. Ma, A.~Ambikapathi, and C.-Y. Chi}, {\em A simplex
  volume maximization framework for hyperspectral endmember extraction}, IEEE
  Trans Geosci Remote Sens, 49 (2011), pp.~4177--4193.

\bibitem{coifman2006diffusion}
{\sc R.~R. Coifman and S.~Lafon}, {\em Diffusion maps}, Appl Comput Harm Anal,
  21 (2006), pp.~5--30.

\bibitem{eismann2012hyperspectral}
{\sc M.~T. Eismann}, {\em Hyperspectral remote sensing}, SPIE, 2012.

\bibitem{friedman2001elements}
{\sc J.~Friedman, T.~Hastie, and R.~Tibshirani}, {\em The elements of
  statistical learning}, Springer Series in Satistics, 2001.

\bibitem{haut2018active}
{\sc J.~M. Haut, M.~E. Paoletti, J.~Plaza, J.~Li, and A.~Plaza}, {\em Active
  learning with convolutional neural networks for hyperspectral image
  classification using a new \uppercase{B}ayesian approach}, IEEE Trans Geosci
  Remote Sens, 56 (2018), pp.~6440--6461.

\bibitem{maggioni2019LAND}
{\sc M.~Maggioni and J.~M. Murphy}, {\em Learning by active nonlinear
  diffusion}, Found Data Sci, 1 (2019), p.~271.

\bibitem{murphy2020spatially}
{\sc J.~M. Murphy}, {\em Spatially regularized active diffusion learning for
  high-dimensional images}, Pattern Recognit Lett, 135 (2020), pp.~213--220.

\bibitem{murphy2018unsupervised}
{\sc J.~M. Murphy and M.~Maggioni}, {\em Unsupervised clustering and active
  learning of hyperspectral images with nonlinear diffusion}, IEEE Trans Geosci
  Remote Sens, 57 (2018), pp.~1829--1845.

\bibitem{murphy2019spectral}
{\sc J.~M. Murphy and M.~Maggioni}, {\em Spectral--spatial diffusion geometry
  for hyperspectral image clustering}, IEEE Geosci Remote Sens Lett, 17 (2019),
  pp.~1243--1247.

\bibitem{murphy2021multiscale}
{\sc J.~M. Murphy and S.~L. Polk}, {\em A multiscale environment for learning
  by diffusion}, Appl Comput Harmon Anal, 57 (2022), pp.~58--100.

\bibitem{nascimento2005VCA}
{\sc J.~M.~P. Nascimento and J.~M. Bioucas-Dias}, {\em Vertex component
  analysis: A fast algorithm to unmix hyperspectral data}, IEEE Trans Geosci
  Remote Sens, 43 (2005), pp.~898--910.

\bibitem{DVIS}
{\sc S.~L. Polk, K.~Cui, R.~J. Plemmons, and J.~M. Murphy}, {\em Diffusion and
  volume maximization-based clustering of highly mixed hyperspectral images},
  arXiv preprint arXiv:2203.09992,  (2022).

\bibitem{polk2021multiscale}
{\sc S.~L. Polk and J.~M. Murphy}, {\em Multiscale clustering of hyperspectral
  images through spectral-spatial diffusion geometry}, in Proc IEEE Geosci
  Remote Sens Symp, 2021, pp.~4688--4691.

\bibitem{tuia2011survey}
{\sc D.~Tuia, M.~Volpi, L.~Copa, M.~Kanevski, and J.~Munoz-Mari}, {\em A survey
  of active learning algorithms for supervised remote sensing image
  classification}, IEEE J Sel Top Appl Earth Obs Remote Sens, 5 (2011),
  pp.~606--617.

\bibitem{wang2017novel}
{\sc Z.~Wang, B.~Du, L.~Zhang, L.~Zhang, and X.~Jia}, {\em A novel
  semisupervised active-learning algorithm for hyperspectral image
  classification}, IEEE Trans Geosci Remote Sens, 55 (2017), pp.~3071--3083.

\end{thebibliography}

\end{document}